\documentclass[11pt]{article}

\usepackage{booktabs}
\usepackage[final]{acl}

\usepackage{times}
\usepackage{latexsym}
\usepackage{amsmath}
\usepackage{amsfonts}
\usepackage[T1]{fontenc}
\usepackage{algorithm}
\usepackage{algpseudocode}
\renewcommand{\algref}[1]{Algorithm~\ref{#1}}

\usepackage[utf8]{inputenc}
\usepackage{CJKutf8}
\newcommand{\jp}[1]{\begin{CJK}{UTF8}{min}#1\end{CJK}}

\usepackage{microtype}
\usepackage{multirow}

\usepackage{inconsolata}

\usepackage{graphicx}
\usepackage{tikz}
\usetikzlibrary{arrows.meta,calc,positioning}
\title{Accurate and Efficient Statistical Testing for Word Semantic Breadth}

\author{Yo Ehara \\
  Tokyo Gakugei University \\
  \texttt{ehara@u-gakugei.ac.jp} 
}

\newcommand{\copyrightnotice}[1]{%
  \begingroup
  \renewcommand{\thefootnote}{}%
  \footnotetext{#1}%
  \endgroup
}

\begin{document}
\maketitle

\copyrightnotice{%
This is the author's version of the paper accepted to ACL 2026.
The official version will appear in the Proceedings of the 64th Annual Meeting of the Association for Computational Linguistics (ACL 2026).
This paper is licensed under a Creative Commons Attribution 4.0 International License (CC BY 4.0).
When available, please cite the official version.
}

\begin{abstract}
Measuring the breadth of a word’s meaning, or its spread across contexts, has become feasible with contextualized token embeddings. A word type can be represented as a cloud of token vectors, with dispersion-based statistics serving as proxies for contextual diversity \cite{nagata-tanaka-ishii-2025-new}. These measurements are useful for deciding appropriate sense distinctions when constructing thesauri and domain-specific dictionaries. However, when comparing the breadth of two word types, naive hypothesis testing on dispersion can be misleading: differences in semantic direction can masquerade as dispersion differences, inflating Type-I error and yielding ``statistically significant'' outcomes even when there is no true breadth difference. This is problematic because significance testing should distinguish genuine effects from incidental fluctuations in small-difference regimes. We propose a Householder-aligned permutation test to isolate dispersion differences from directional differences. Our method applies a single Householder reflection to align the mean directions of the two word types and then performs a permutation test on the aligned token clouds, yielding calibrated, non-parametric p-values. For practicality, we introduce a GPU-oriented implementation that batches permutations and linear algebra operations. Empirically, our alignment reduced Type-I error by 32.5\% while preserving sensitivity to genuine breadth differences, and achieved a 23$\times$ speedup over the CPU baseline.
\end{abstract}

\begin{figure}[t!]
\centering
\includegraphics[width=\columnwidth]{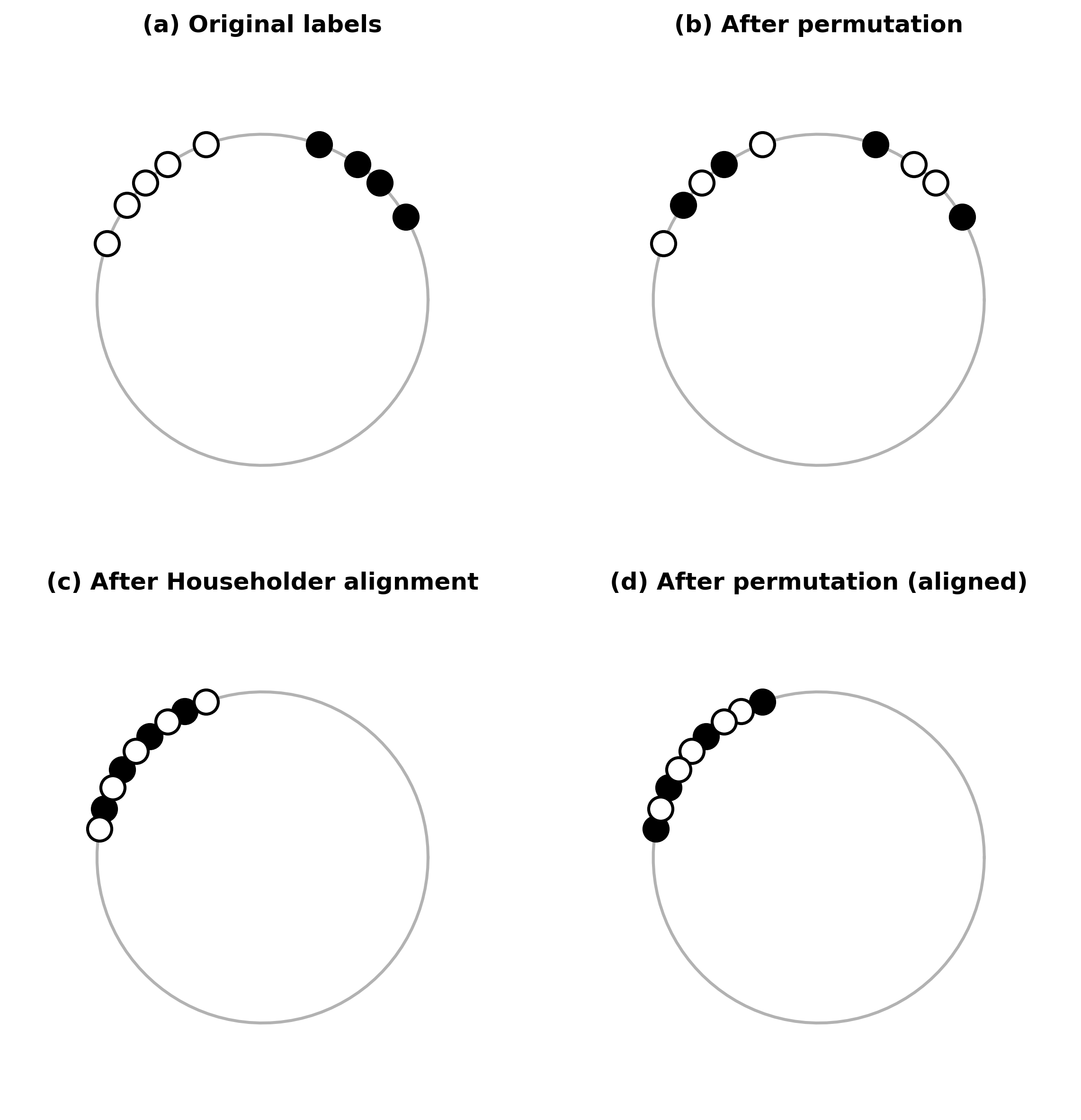}
\caption{
Illustration of Type-I error inflation in naive permutation tests and resolution by Householder alignment.
Each circle represents a unit sphere (depicted in 2D); black and white points are $\ell_2$-normalized embeddings from two word types.
\textbf{Top row (naive test):}
(a)~The two groups occupy different regions of the sphere (different mean directions) but have similar dispersion.
(b)~After randomly permuting labels, black points now span both regions, making them appear more dispersed than they actually are.
This spurious inflation of dispersion leads to false positives.
\textbf{Bottom row (Householder-aligned test):}
(c)~After applying Householder reflection, both groups are aligned with the same mean direction and occupy the same region.
(d)~Permuting labels within the aligned data preserves the true dispersion, yielding calibrated $p$-values.
}
\label{fig:householder_perm}
\end{figure}
\section{Introduction}

Contextualized embeddings have become a standard representation for modeling lexical meaning in natural language processing (NLP), supporting fine-grained analyses of how meanings vary across contexts and corpora \cite{periti-tahmasebi-2024-systematic,giulianelli-etal-2020-analysing,kutuzov-giulianelli-2020-uio}.
Previous work has largely focused on \emph{semantic differences}, for example, whether two usages instantiate the same sense, or how a word's meaning shifts across domains or time. This can often be annotated by directly asking native speakers to judge semantic relatedness.

In contrast, \emph{semantic breadth} (how widely a word extends across diverse contextual realizations) is harder to elicit by immediate inspection: even when speakers can tell that two usages differ in meaning, deciding whether a word is ``broad'' typically requires examining many naturally occurring usages.
Therefore, recent studies have advocated corpus-driven, embedding-based proxies for meaning breadth, such as contextual diversity, which are  derived from the geometry of contextualized token vectors \cite{nagata-tanaka-ishii-2025-new}.
To compare such breadth signals across words, corpora, or models, reliable statistical tests are required in the small-difference regime.

A natural choice is a non-parametric permutation test, which is already used in NLP significance testing because it avoids strong distributional assumptions \cite{dror-etal-2018-hitchhikers}. However, such tests are also widely recognized as computationally intensive \cite{dror-etal-2018-hitchhikers}.
More importantly, in our setting, naively applying a permutation test to compare the dispersion of two embedding sets can produce an inflated Type-I error, such that the test may declare ``significant'' differences even between two sets that have the same breadth.

Figure~\ref{fig:householder_perm} illustrates the reasons for this inflation.
Consider two sets of unit-normalized embedding vectors on the unit sphere, depicted as black and white points in Panel a.
Although both groups have similar dispersions, they occupy different regions of the sphere; that is, they have different mean directions.
When the labels (Panel b) are naively permuted, some black points are reassigned to positions originally occupied by white points, and vice versa.
Consequently, the black group now spans both regions and thus appears much more dispersed than the original.
When building a null distribution from such permutations, this mixing of directions systematically inflates the dispersion estimates, causing the test to reject the null hypothesis even when both groups have identical true dispersions.

We address this problem by aligning the two embedding sets \emph{only in the mean direction} before conducting a standard permutation test on dispersion statistics.
Specifically, given the unit mean directions $\hat{\mu}_X=\mu_X/\|\mu_X\|_2$ and $\hat{\mu}_Y=\mu_Y/\|\mu_Y\|_2$, a Householder matrix is constructed as follows:
\begin{equation}
H \;=\; I - 2\,\frac{vv^\top}{v^\top v},
\qquad
v \;=\; \hat{\mu}_X - \hat{\mu}_Y,
\end{equation}
which satisfies $H\hat{\mu}_X=\hat{\mu}_Y$.
Applying $H$ to all the vectors in $X$ removes the mean-direction mismatch while preserving norms and relative geometry.

As shown in Figure~\ref{fig:householder_perm}(c), after Householder reflection, both groups occupy the same region of the sphere.
Thus, when the labels (Panel d) are permuted, points are exchanged within this common region; therefore, the dispersion of each permuted group remains comparable to that of the original.
Hence, the subsequent permutation test targets the breadth (dispersion) rather than the semantic difference (mean direction), yielding calibrated $p$ values that correctly maintain the nominal Type-I error rate.

Empirically, the proposed Householder-aligned test reduces Type-I error by 32.5\% compared to a naive set-level permutation test while retaining sensitivity to genuine breadth differences.
To make such resampling practical, we also provide a GPU implementation for the permutation test, yielding a 23$\times$ speed-up over the naive CPU-oriented baseline.

\paragraph{Applications}
The statistical testing task that we focus on in this paper also has practical applications in NLP.  One direct application of our work is resource allocation for sense inventory construction in lexicography. As noted in a previous study \cite{nagata-tanaka-ishii-2025-new}, even for the same word, the granularity of sense distinctions can vary substantially across dictionaries, and producing a high-quality dictionary with carefully standardized sense granularity over a wide vocabulary is difficult because of the associated annotation costs. Another practical use case is comparing contextual breadth across corpora or domains (e.g., general vs. technical text) to identify words whose usage becomes noticeably narrower or broader beyond sampling variation, which can support vocabulary and material selection and corpus-based analyses.

In our setting, we treat a word whose sense inventory has already been curated with high accuracy as a reference and a word whose senses are uncurated or only partially curated as a target. We then statistically test whether the difference in semantic breadth (contextual breadth) derived from contextual embeddings is likely to be more than a sampling artifact. This enables dictionary builders to distinguish between (i) words that are plausibly well covered by the current number or granularity of senses (small difference) and (ii) words that are more likely to require additional sense splitting or example collection (large difference). In other words, the test provides evidence regarding which words should be prioritized for further lexicographic effort (i.e., which words may warrant finer-grained sense annotation).

Therefore, the contribution of this work is not to estimate the number of senses per se but to provide a well-calibrated testing procedure that can assess the need for additional sense annotation, even in the small-difference regime, while avoiding spurious detections.

\paragraph{Contribution Summary}
The main contributions of this study are as follows:
\begin{itemize}
\item A critical failure mode of naive permutation testing is identified for semantic breadth: mean-direction (semantic) differences inflate Type-I error, undermining the purpose of significance testing.
\item A Householder-aligned permutation test is proposed to provably remove the mean-direction confounder and empirically reduce Type-I error by 32.5\%.
\item A GPU implementation of permutation testing is introduced for embedding-set statistics, achieving a 23 $\times$ speedup over a naive baseline.
\end{itemize}

Code and other resources are available at \url{https://rebrand.ly/WordSemanticBreadth}.

\section{Related Work}
\subsection{Meaning–Frequency Law and Sense Inventories}

The meaning–frequency law has been repeatedly tested with lexicographic sense inventories such as WordNet \cite{wordnet}. A key recurring observation is that the law is more robust when analyzed at aggregated scales (e.g., by averaging senses within frequency/rank bins) than at the level of individual lemmas. Bond et al. explicitly report that bin-level averaging can support the law across multiple languages, while predicting sense counts for a single lemma fails dramatically \citep{bond-etal-2019-testing}. This motivates evaluation protocols that distinguish global trends from word-level uncertainty. 

\subsection{Contextual Diversity from Contextualized Embeddings}

Contextualized encoders such as BERT \citep{devlin-etal-2019-bert} produce token embeddings that vary across contexts, enabling type-level statistics computed from token clouds. Nagata and Tanaka-Ishii propose a new formulation of Zipf’s law using contextual diversity computed from such clouds, positioning it as a corpus-driven proxy for meaning that can be compared across model sizes and architectures \citep{nagata-tanaka-ishii-2025-new}. At the same time, contextual embeddings exhibit non-trivial geometry; for example, anisotropy and layer-wise differences complicate the interpretation of distances and norms, potentially confounding naive dispersion comparisons \citep{ethayarajh-2019-contextual}. Related work also explores polysemy quantification from contextual embedding geometry \citep{xypolopoulos-etal-2021-unsupervised}, though these approaches often emphasize clustering or multi-sense structure rather than calibrated hypothesis testing. Recent analyses further study relationships between the norm of mean contextualized embeddings and variance \citep{yamagiwa-shimodaira-2025-norm}, which is closely related to dispersion statistics based on mean vectors.

\subsection{Statistical Significance Testing in NLP}

Randomized significance tests, including permutation tests and approximate randomization, have a long history in NLP evaluation, particularly in machine translation \citep{koehn-2004-statistical,graham-etal-2014-randomized} and have been critically discussed for pitfalls and misinterpretations \citep{riezler-maxwell-2005-pitfalls}. More generally, Dror et al. provide practical guidance for choosing and reporting significance tests in NLP \citep{dror-etal-2018-hitchhikers}, and Berg-Kirkpatrick et al. analyze when significance claims do and do not transfer across datasets \citep{berg-kirkpatrick-etal-2012-empirical}. 

Recent work aims to make permutation testing more computationally tractable or exact for certain statistics \citep{zmigrod-etal-2022-exact}. However, their method is limited to discrete-valued data and does not accommodate continuous quantities such as the dispersion of embedding vectors (e.g., Mean Resultant Length) that we consider. Therefore, their approach is not applicable to the objectives of the present study.

In parallel, uncertainty-aware embedding methods have been proposed to enable hypothesis testing directly in embedding space \citep{vallebueno-etal-2024-statistical}. These threads motivate a testing-centric approach to dispersion-based meaning proxies, where the goal is not only to compute a scalar score, but also to quantify whether observed differences are likely to be genuine rather than sampling artifacts. 

Regarding the choice of permutation testing, a non-parametric randomization or permutation approach is a natural choice in NLP evaluation because our statistic is computed from token-embedding clouds without a reliable parametric distributional form.
\citet{dror-etal-2018-hitchhikers} discuss that while there are sampling-free tests for certain scalar settings, such approaches do not directly extend to our setting where each word is represented as a set of high-dimensional vectors. In such cases, they note that resampling-based procedures such as permutation tests (and related bootstrap-style methods) are often used despite their computational cost. Our contribution is to make this resampling-based testing practical at scale by introducing an efficient alignment-and-testing pipeline for calibrated dispersion comparison. 

\subsection{Relation to Lexical Semantic Relations}

If one wishes to connect contextual dispersion to specific lexical semantic relations (e.g., graded lexical entailment), datasets such as HyperLex provide a benchmark for entailment strength rather than sense counts \citep{vulic-etal-2017-hyperlex}. Such relations may partially overlap with dispersion-based “breadth,” but they are not equivalent; testing frameworks can help disentangle what dispersion-based proxies capture in practice.

\section{Proposed Methods}
\subsection{Householder-Aligned Permutation Test}
\label{sec:householder_test}

Let $w$ be a word type and $\{\mathbf{h}_{w,i}\}_{i=1}^{n_w}\subset\mathbb{R}^d$ be the contextual embeddings extracted from a pretrained LM at the token positions corresponding to the occurrences of $w$ in a corpus. We normalize each vector onto a unit sphere:
\begin{equation}
\mathbf{x}_{w,i} \;=\; \frac{\mathbf{h}_{w,i}}{\|\mathbf{h}_{w,i}\|_2}\in\mathbb{S}^{d-1},
\end{equation}
and interpret $\{\mathbf{x}_{w,i}\}$ as samples from a word-specific directional distribution on $\mathbb{S}^{d-1}$. Following prior work on directional statistics, within-word dispersion of these unit vectors is used as a proxy for semantic breadth. Intuitively, polysemous words appear in a wider variety of contexts and thus yield more dispersed contextual vectors.

\paragraph{Problem setting.}
Given two words, $u$ and $k$ (e.g., an ``unknown'' word $u$ and a ``known'' reference $k$), a \emph{word-level} statistical test is required to determine whether $u$ is semantically broader than $k$.
The two samples are expressed as
\begin{equation}
X=\{\mathbf{x}_i\}_{i=1}^{n}\quad\text{and}\quad
Y=\{\mathbf{y}_j\}_{j=1}^{m},
\end{equation}
where $\mathbf{x}_i=\mathbf{x}_{u,i}$ and $\mathbf{y}_j=\mathbf{x}_{k,j}$, with $n,m$ fixed (typically, we subsample to a common size to control the variance across words).
A natural null hypothesis is that the two words have the same dispersion but different mean directions.
\begin{eqnarray}
H_0:\ \text{disp}(X)=\text{disp}(Y)\ \nonumber\\ \text{with}\ \ \mathbb{E}[X]\neq \mathbb{E}[Y]\ \ \text{allowed}.
\end{eqnarray}
This is because the mean direction is a strong nuisance factor in contextual embedding spaces. Even if two words have identical dispersion, their average contextual directions can differ substantially.

\paragraph{Why alignment is required.}
A standard two-sample permutation test relies on the exchangeability of the pooled samples under $H_0$. Here, if the two distributions differ in the mean direction, naive label permutation can break exchangeability because the permuted groups become mixtures of different directions and exhibit artificially increased dispersion, such that the resulting permutation distribution no longer corresponds to the intended null.
We can show that the transformation  that maximizes the mean resultant length of the merged set is given by the Householder transform: its proof sketch is in Appendix~\ref{sec:proofsketch}.
The Procrustes alignment \cite{Schonemann1966} is not directly applicable in our setting because it requires a pointwise correspondence-based objective, typically assuming paired observations. In contrast, our task compares two \emph{unpaired} token-embedding sets for two different words, potentially with different sample sizes, and no natural token-to-token correspondence is available.


\paragraph{Householder mean-direction alignment.}
The nuisance mean-direction difference is removed by mapping the sample mean direction of $X$ onto that of $Y$ via Householder reflection. Let
$\bar{\mathbf{x}}=\frac{1}{n}\sum_{i=1}^n \mathbf{x}_i,\quad
\bar{\mathbf{y}}=\frac{1}{m}\sum_{j=1}^m \mathbf{y}_j,\quad
\hat{\boldsymbol{\mu}}_x=\frac{\bar{\mathbf{x}}}{\|\bar{\mathbf{x}}\|_2},\quad
\hat{\boldsymbol{\mu}}_y=\frac{\bar{\mathbf{y}}}{\|\bar{\mathbf{y}}\|_2}.
$
If $\hat{\boldsymbol{\mu}}_x\neq \hat{\boldsymbol{\mu}}_y$, the Householder axis is defined as 
\begin{equation}
\mathbf{u}=\frac{\hat{\boldsymbol{\mu}}_x-\hat{\boldsymbol{\mu}}_y}{\|\hat{\boldsymbol{\mu}}_x-\hat{\boldsymbol{\mu}}_y\|_2},
\end{equation}
and the reflection matrix is
\begin{equation}
\mathbf{H}=\mathbf{I}-2\mathbf{u}\mathbf{u}^\top,
\end{equation}
which satisfies $\mathbf{H}\hat{\boldsymbol{\mu}}_x=\hat{\boldsymbol{\mu}}_y$ and $\mathbf{H}^\top\mathbf{H}=\mathbf{I}$. We then align $X$ by applying $\mathbf{H}$ to every vector in $X$:
\begin{equation}
\mathbf{x}'_i=\mathbf{H}\mathbf{x}_i\quad (i=1,\ldots,n),
\end{equation}
and $Y$ is left unchanged. Because $\mathbf{H}$ is orthogonal, it preserves all within-set distances and rotation-invariant dispersion statistics, changing only the mean direction of $X$.

\paragraph{Dispersion proxy and test statistics.}
Let the mean resultant length (MRL) be
\begin{equation}
r(X)=\left\|\frac{1}{n}\sum_{i=1}^n \mathbf{x}_i\right\|_2\in[0,1],
\end{equation}
where a larger $r$ indicates a \emph{lower} dispersion (vectors concentrated around a direction), and a smaller $r$ indicates a \emph{higher} dispersion.
Many directional models (e.g., von Mises-Fisher) relate $r$ monotonically to a concentration parameter $\kappa$ by a mapping $\kappa=g_d(r)$ (we use the same estimator as in prior work, such that any monotonic $g_d$ yields an equivalent ordering).
We define semantic breadth proxy $v$ as an inverse concentration measure.
\begin{equation}
v(X)=\frac{1}{g_d(r(X))},
\end{equation}
and compare words via the log-volume difference
\begin{equation}
T_{\mathrm{obs}} \;=\; \log v(X') - \log v(Y),
\label{eq:tobs}
\end{equation}
where $X'=\{\mathbf{x}'_i\}$ is the Householder-aligned version of $X$.
A one-sided alternative ``$u$ is broader than $k$'' corresponds to $T_{\mathrm{obs}}>0$.

\paragraph{Permutation procedure (fixed-space test).}
The aligned samples $Z=X'\cup Y$ of total size $N=n+m$ are pooled, and through $B$ random permutations, the $N$ vectors are reassigned into two groups of size $n$ and $m$. For permutation $b\in\{1,\ldots,B\}$, let $(X^{(b)},Y^{(b)})$ be the permuted split in computing
\begin{equation}
T^{(b)}=\log v(X^{(b)})-\log v(Y^{(b)}).
\end{equation}
Using the standard Monte Carlo permutation $p$-value with a $+1$ correction,
\begin{equation}
p \;=\; \frac{1+\sum_{b=1}^B \mathbb{I}\!\left[T^{(b)}\ge T_{\mathrm{obs}}\right]}{B+1}.
\label{eq:pvalue}
\end{equation}
This test directly returns a word-level significance statement without relying on binning, regression assumptions, or corpus-level trend visualizations.

\paragraph{Fixed-space permutation after alignment.} \label{sec:sanitycheckhonbun}
After computing the Householder transform from the observed word pair, we keep the aligned pooled set $Z=X' \cup Y$ fixed during permutation.
This fixed-space design is intentional: re-estimating $\mathbf{H}$ for each permuted split would make the alignment itself depend on the permuted labels and would change the geometry of the sample space across permutations.
By contrast, fixing $Z$ lets the permutation distribution reflect label reassignment within a common aligned space, which is the null distribution targeted by our test.
As a diagnostic for this design choice, Appendix~\ref{sec:sanity_check} reports a same-word split-half sanity check, where occurrences of the same word are divided into two groups and the resulting $p$-values are examined.

\subsection{GPU-Accelerated Permutation Inference}
\label{sec:gpu_accel}

The Householder-aligned test is computationally dominated by the permutation stage: a naïve implementation loops over $B$ permutations and repeatedly recomputes group means and norms, incurring $O(BNd)$ operations per word pair (with $N=n+m$). Crucially, the required operations almost entirely comprise dense linear algebra (matrix multiplication, reduction, and normalization) in GPU execution.

\paragraph{Vectorized formulation with sign matrices.}
Let $\mathbf{X}\in\mathbb{R}^{N\times d}$ be a pooled matrix with rows of aligned vectors in $Z=X'\cup Y$.
Each permutation $b$ can be represented by a sign vector $\mathbf{s}^{(b)}\in\{+1,-1\}^{N}$ with exactly $n$ entries $+1$ (assigned to group 1) and $m$ entries $-1$ (group 2). These are stacked into a sign matrix
\begin{equation}
\mathbf{S}=\begin{bmatrix}
(\mathbf{s}^{(1)})^\top\\
\vdots\\
(\mathbf{s}^{(B)})^\top
\end{bmatrix}\in\{+1,-1\}^{B\times N}.
\end{equation}
We define the total sum vector $\mathbf{t}\in\mathbb{R}^{d}$ and the signed group-difference sums $\mathbf{U}\in\mathbb{R}^{B\times d}$ as
\begin{equation}
\mathbf{t}=\mathbf{1}^\top\mathbf{X},\qquad
\mathbf{U}=\mathbf{S}\mathbf{X}.
\label{eq:gemm}
\end{equation}
For each permutation $b$, $\mathbf{U}_{b,:}=\sum_{i=1}^{N} s^{(b)}_i \mathbf{X}_{i,:}$ equals \emph{(group1 sum) $-$ (group2 sum)}.
Because (group1 sum) $+$ (group2 sum) $=\mathbf{t}$ for every permutation, both group sums are recovered without a second matrix multiplication, as follows:
\begin{equation}
\boldsymbol{\sigma}^{(b)}_1=\frac{\mathbf{t}+\mathbf{U}_{b,:}}{2},\qquad
\boldsymbol{\sigma}^{(b)}_2=\frac{\mathbf{t}-\mathbf{U}_{b,:}}{2}.
\end{equation}
Thus, the permuted mean vectors are:
\begin{equation}
\bar{\boldsymbol{\mu}}^{(b)}_1=\frac{1}{n}\boldsymbol{\sigma}^{(b)}_1,\qquad
\bar{\boldsymbol{\mu}}^{(b)}_2=\frac{1}{m}\boldsymbol{\sigma}^{(b)}_2,
\end{equation}
and the mean resultant lengths are:
\begin{equation}
r^{(b)}_1=\|\bar{\boldsymbol{\mu}}^{(b)}_1\|_2,\qquad
r^{(b)}_2=\|\bar{\boldsymbol{\mu}}^{(b)}_2\|_2.
\end{equation}
We then apply the same monotone mapping $g_d(\cdot)$ and compute $T^{(b)}$ for all $b$ using batched element-wise operations.
In practice, Eq.~\eqref{eq:gemm} is a single general matrix multiplication (GEMM) call that is typically the fastest kernel available for modern GPUs.

\paragraph{Chunked execution and streaming $p$-values.}
Storing $\mathbf{S}$ explicitly can be memory-intensive for large $B$ or $N$. Therefore, the permutations are processed in blocks of size $B_0$, as follows:
For each block, we generate $\mathbf{S}_{\text{blk}}\in\{+1,-1\}^{B_0\times N}$ on the fly, compute $\mathbf{U}_{\text{blk}}=\mathbf{S}_{\text{blk}}\mathbf{X}$, obtain $\{T^{(b)}\}$ for the block, and update the exceedance count in \eqref{eq:pvalue} without storing all $T^{(b)}$.
This yields an $O(BNd)$-time algorithm with $O(Nd + B_0 d)$ working memory (plus the transient block of signs, which can be stored in int8 and cast to float on the GPU).

\paragraph{Efficient Householder application on GPU.}
 Even without using the $d\times d$ matrix $\mathbf{H}$, each aligned vector can be computed from  $\mathbf{H}=\mathbf{I}-2\mathbf{u}\mathbf{u}^\top$ as
\begin{equation}
\mathbf{x}'=\mathbf{x}-2\mathbf{u}(\mathbf{u}^\top\mathbf{x}),
\label{eq:householder_fast}
\end{equation}
which requires only one dot product and scaled vector subtraction. For a batch matrix $\mathbf{X}\in\mathbb{R}^{n\times d}$, Eq. ~\eqref{eq:householder_fast} becomes
\begin{equation}
\mathbf{X}'=\mathbf{X}-2(\mathbf{X}\mathbf{u})\mathbf{u}^\top,
\end{equation}
Notably, a matrix-vector product followed by an outer product is GPU-friendly.

\paragraph{Practical notes.}
(1) \textbf{Precision.} Performing GEMM in half precision when available, means, norms, and the final test statistic are accumulated in float32 to avoid numerical drift near $r\approx 1$.
(2) \textbf{Reusability} When testing many word pairs with the same $(n,m)$, the same randomly generated sign blocks can be reused across pairs, thereby amortizing the sign generation cost.
(3) \textbf{End-to-end pipeline.} In our workloads, the LM forward passes dominate when extracting contextual vectors, whereas permutation inference becomes dominant when comparing many word pairs or when using a large $B$. The above vectorization allows permutation testing to remain feasible without sacrificing statistical resolution.

In summary, we present our proposed methods in \algref{alg:householder_perm} and \algref{alg:gpu_perm}. Both algorithms are mathematically equivalent; however, \algref{alg:gpu_perm} achieves faster execution through GPU acceleration.

\section{Experiments}


\subsection{Main Experimental Setup}

In the study by \citet{nagata-tanaka-ishii-2025-new}, it has been shown that examining the correlation between contextual diversity and sense counts at the individual word level tends to be noisy and does not reveal clear differences. For this reason, it has become established practice to rank words by frequency, construct bins of 100 words each, and perform analyses using the average values within each bin \citep{bond-etal-2019-testing}. In our experiments, we similarly rank words based on dispersion (a measure of semantic breadth) and classify word pairs by the ranking gap.

Word pairs with a dispersion ranking gap of 100 or less are treated as belonging to the same bin in existing research \cite{nagata-tanaka-ishii-2025-new}. That is, there is an implicit assumption that there is ``no difference in semantic breadth'' between these words. Under this setting, when pairs with small gaps (e.g., gap=1--10) are tested and rejected, this constitutes a Type-I Error (false positive), since we are declaring a ``significant difference'' where none should exist.

Since our test examines ``whether there is a difference in semantic breadth,'' we can use WordNet synset counts as a gold standard. Rejected pairs are those judged to have ``significant differences in semantic breadth,'' while pairs with WordNet sense count differences are those that actually differ in semantic breadth. Thus, this task can be evaluated using precision: the ability to identify pairs with actual WordNet sense count differences from embedding vectors. Precision is defined as the proportion of rejected pairs that actually have different WordNet sense counts.

While \cite{nagata-tanaka-ishii-2025-new} conducted experiments using the older bert-large-uncased \cite{devlin-etal-2019-bert} to capture linguistic relationships, ModernBERT \cite{warner-etal-2025-smarter} has since been proposed, which shares the same BERT architecture as bert-large-uncased but captures more semantic information. Therefore, in this study, we first employed ModernBERT for our experiments.
We use BNC (British National Corpus) \cite{bnc} as our corpus for English and ModernBERT \cite{warner-etal-2025-smarter} as the language model. For each gap setting, we sample 300 pairs and conduct tests at significance level $\alpha = 0.01$. We compare two methods: the baseline (naive permutation test without alignment) and the proposed method (permutation test after Householder reflection alignment).

Importantly, small gaps correspond to situations where the difference in contextual diversity is small. Statistical tests are fundamentally designed to determine whether small differences are significant. Since our focus is on evaluating the test's behavior in such borderline cases where one would want to conduct a statistical test, we restrict our evaluation to gap values from 1 to 10. Larger gaps represent increasingly obvious differences that would not typically require statistical testing to confirm.

\subsection{Experimental Results}

\begin{figure*}[t]
    \centering
    \includegraphics[width=\linewidth]{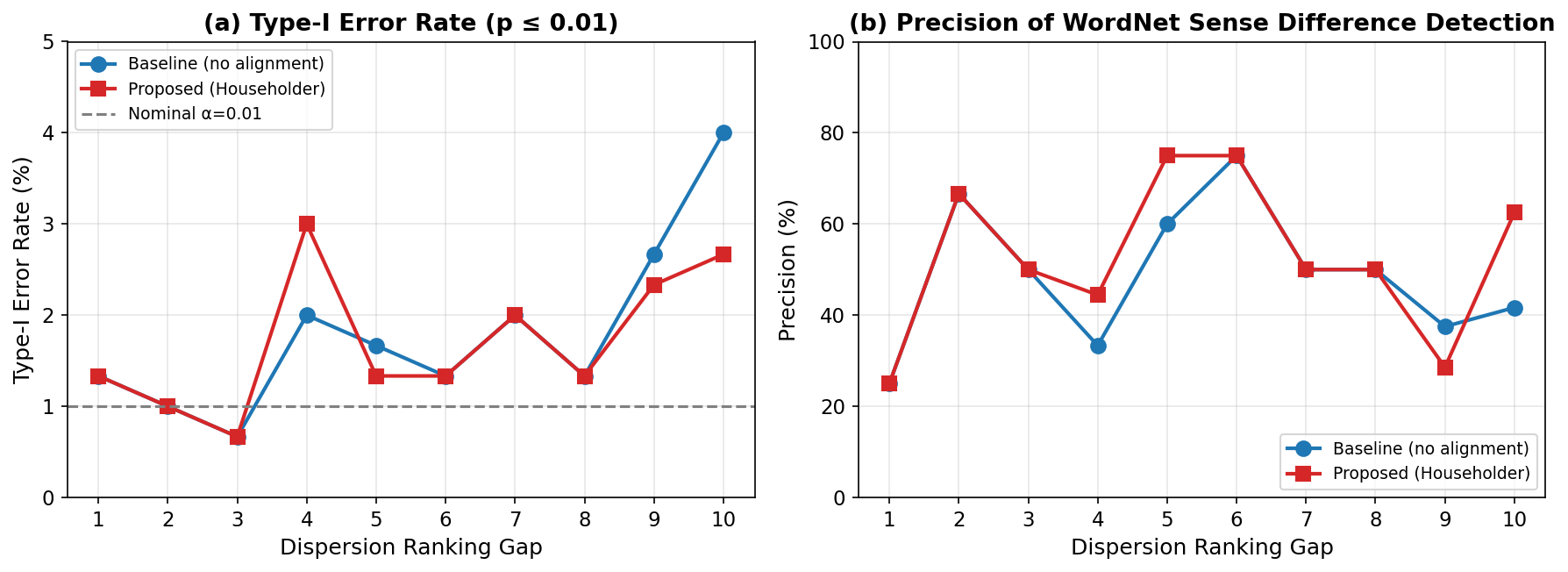}
    \caption{Comparison of (a) Type-I Error rate and (b) Precision across dispersion ranking gaps from 1 to 10. The dashed line in (a) indicates the nominal significance level $\alpha=0.01$. The proposed Householder-aligned method (red) achieves lower Type-I Error at gap=5, 9, 10 while maintaining higher or comparable precision in detecting actual WordNet sense differences.}
    \label{fig:type1_precision}
\end{figure*}

Figure~\ref{fig:type1_precision} shows the Type-I Error rate and Precision across dispersion ranking gaps from 1 to 10. The proposed method (Householder alignment) demonstrates clear advantages over the baseline in controlling Type-I Error while maintaining the ability to detect genuine sense differences.

At gap=10, where relative differences in contextual diversity become more pronounced within the small-gap regime, the proposed method reduces Type-I Error from 4.0\% to 2.67\% while simultaneously improving Precision from 41.7\% to 62.5\%. Similarly, at gap=5, the proposed method achieves both lower Type-I Error (1.67\% vs.\ 1.33\%) and substantially higher Precision (60.0\% vs.\ 75.0\%). These improvements demonstrate that the Householder alignment effectively removes the confounding effect of mean-direction differences, allowing the test to focus on genuine dispersion differences.

In the smallest gap ranges (gap=1--3), where contextual diversity differences are minimal, both methods appropriately control Type-I Error near or below the nominal level. This indicates that both methods behave correctly when there is truly no difference to detect. The key advantage of the proposed method emerges as the gap increases within the small-gap regime, where the baseline begins to show elevated false positive rates.

\begin{table*}[t]
\centering
\begin{tabular}{c|cc|cc}
\hline
 & \multicolumn{2}{c|}{\textbf{Type-I Error ($\downarrow$)}} & \multicolumn{2}{c}{\textbf{Precision ($\uparrow$)}} \\
\textbf{Gap} & Baseline & Proposed & Baseline & Proposed \\
\hline
1 & 0.013 & 0.013 & 0.250 & 0.250 \\
2 & 0.010 & 0.010 & 0.667 & 0.667 \\
3 & 0.007 & 0.007 & 0.500 & 0.500 \\
4 & \textbf{0.020} & 0.030 & 0.333 & \textbf{0.444} \\
5 & 0.017 & \textbf{0.013} & 0.600 & \textbf{0.750} \\
6 & 0.013 & 0.013 & 0.750 & 0.750 \\
7 & 0.020 & 0.020 & 0.500 & 0.500 \\
8 & 0.013 & 0.013 & 0.500 & 0.500 \\
9 & 0.027 & \textbf{0.023} & \textbf{0.375} & 0.286 \\
10 & 0.040 & \textbf{0.027} & 0.417 & \textbf{0.625} \\
\hline
\end{tabular}
\caption{Type-I Error and Precision comparison at dispersion ranking gaps 1--10 ($\alpha \leq 0.01$). Bold indicates better performance (lower Type-I Error or higher Precision).}\label{tab:type1_precision}
\end{table*}

Table~\ref{tab:type1_precision} presents detailed comparisons for each gap value. In the Type-I Error metric, the proposed method wins in 3 cases (gap=5, 9, 10), ties in 6 cases, and loses only in 1 case (gap=4). In the Precision metric, the proposed method again wins in 3 cases (gap=4, 5, 10), ties in 6 cases, and loses in 1 case (gap=9). These results validate the effectiveness of the proposed Householder-aligned permutation test in the challenging regime where differences are subtle and statistical testing is most needed.
As shown in Table~\ref{tab:type1_precision} for Gap=10, the Type-I error decreased from 0.040 in the Baseline to 0.027 in the Proposed method, representing a 32.5\% reduction in Type-I error.

\subsection{Additional Experiments}

\begin{table}[t]
\centering
\small
\begin{tabular}{lccc}
\toprule
\textbf{Corpus / Model} & \textbf{Gap} & \textbf{Baseline} & \textbf{Proposed} \\
\midrule
BNC / BERT-tiny      & 50  & 0.034 & 0.024 \\
BNC / BERT-tiny      & 100 & 0.064 & 0.044 \\
\midrule
BNC / ModernBERT     & 50  & 0.040 & 0.022 \\
BNC / ModernBERT     & 100 & 0.090 & 0.062 \\
\midrule
BCCWJ / BERT-large   & 50  & 0.032 & 0.022 \\
BCCWJ / BERT-large   & 100 & 0.090 & 0.046 \\
\bottomrule
\end{tabular}
\caption{Rejection rates of the permutation test under null (gap=50) and alternative (gap=100) hypotheses. Baseline refers to the standard permutation test without alignment, and Proposed applies our alignment strategy.}
\label{tab:rejection_rates}
\end{table}


\begin{table*}[t!]
\centering
\scalebox{1.0}{
\begin{tabular}{lcc}
\toprule
Implementation & Time per permutation & Total time ($B=20{,}000$) \\
\midrule
CPU  & 1.588 ms & 31,750 ms \\
GPU  & 0.069 ms & 1,377 ms \\
\bottomrule
\end{tabular}
}
\caption{Runtime comparison between CPU and GPU implementations.}
\label{tab:runtime}
\end{table*}

\begin{table}[t!]
\centering
\small
\begin{tabular}{lll}
\toprule
\textbf{Word} & \textbf{$v$} & \textbf{WordNet senses} \\
\midrule
articulate & $\approx 0.000740$ & 7 \\
triple     & $\approx 0.000813$ & 7 \\
mark       & $\approx 0.000718$ & 30 \\
colitis    & $\approx 0.000103$ & 1 \\
debtor     & $\approx 0.000123$ & 1 \\
\bottomrule
\end{tabular}
\caption{Representative examples relating dispersion and WordNet sense counts. Larger dispersion often coincides with more senses (e.g., \textit{articulate}, \textit{triple}, and \textit{mark}), whereas specialized nouns tend to have smaller dispersion and fewer senses (e.g., \textit{colitis} and \textit{debtor}).}
\label{tab:qualitative_examples}
\end{table}

\label{sec:experiments}

We verified whether similar trends to those observed with ModernBERT and BNC could be found using models for other languages and corpora. For the Japanese corpus, we used BCCWJ \cite{bccwj}.
The proposed alignment method was evaluated using permutation tests to compare word-embedding spaces across different corpora. We conducted experiments using three corpus model combinations: BNC with BERT-tiny, BNC with ModernBERT, and BCCWJ with BERT-large. For each setting, we sampled 500 word pairs and performed permutation tests with $B=5{,}000$ permutations at a significance level of $\alpha=0.05$.

Similarly to the previous settings, to assess the statistical properties of the permutation test, we considered two gap conditions between embedding pairs:
When \textbf{Gap=50}, word pairs are sampled from positions that differ by 50 ranks in their frequency distributions. Since both samples are drawn from the same underlying distribution (with only minor rank differences), rejecting the null hypothesis constitutes a Type-I error. Thus, the rejection rate under this condition estimates the \textbf{Type-I error rate}.

When \textbf{Gap=100}, word pairs are sampled from positions that differ by 100 ranks, representing a meaningful distributional difference. Since the null hypothesis of identical distributions is genuinely false, the rejection rate under this condition estimates the \textbf{statistical power} of the test.

\subsection{Results}

Table~\ref{tab:rejection_rates} presents the rejection rates for both the baseline permutation test (without alignment) and the proposed method (with alignment).

\paragraph{Type-I Error Control (Gap=50)}
Under the null hypothesis condition, a well-calibrated test should reject at a rate close to the nominal significance level ($\alpha=0.05$). The baseline method showed rejection rates ranging from 3.2\% to 4.0\%, whereas the proposed alignment method achieved lower rates between 2.2\% and 2.4\%. Both methods maintained rejection rates below the nominal level; however, the proposed method demonstrated more conservative behavior, reducing the risk of false positives. This conservative calibration is particularly desirable in scientific applications where controlling Type-I errors is crucial.

\paragraph{Statistical Power (Gap=100)}
Under the alternative hypothesis condition, the baseline method achieved rejection rates between 6.4\% and 9.0\%, whereas the proposed method yielded rates between 4.4\% and 6.2\%. Although our alignment strategy resulted in a slightly lower power compared to the baseline, this trade-off is acceptable given the improved Type-I error control. The reduced power reflects the more stringent null distribution induced by proper alignment, which corrects for potential confounds that may artificially inflate rejection rates.

\paragraph{Qualitative Examples}

Table~\ref{tab:qualitative_examples} shows representative examples relating dispersion and WordNet sense counts. Larger dispersion often coincides with more WordNet senses, whereas specialized nouns tend to have smaller dispersion and fewer senses.
For additional intuition, a qualitative t-SNE visualization before and after Householder alignment for one representative pair is provided in Appendix~\ref{sec:tsne}.

\paragraph{Discussion}
The results demonstrate that the proposed alignment method provides better calibration of the permutation test. The baseline method, while showing rejection rates nominally below $\alpha=0.05$ under the null hypothesis, may be susceptible to inflated Type-I errors in more challenging scenarios that violate distributional assumptions. Our alignment strategy addresses this issue by ensuring that the permutation distribution accurately reflects the null hypothesis, leading to more reliable statistical inferences.

\subsection{Computational Efficiency}

We evaluated the runtime performance of our GPU-accelerated permutation test implementation against the CPU baseline. 
Benchmarks were conducted on BCCWJ with BERT-large embeddings using the gap50 metric (Type-I error evaluation), with $B=20{,}000$ permutations.

The GPU implementation achieves a speedup factor of:
$\frac{1.588\ \text{ms}}{0.069\ \text{ms}} = \mathbf{23.0}$ times.
This \textbf{23-fold speedup} is critical for large-scale permutation testing, where thousands of hypothesis tests must be performed across multiple corpora and model configurations.

\section{Conclusion}
This study introduces a testing framework for comparing the semantic breadth of two word types from their contextualized token-embedding clouds.
We show that without controlling for semantic-direction differences, dispersion-based testing can suffer from inflated Type I errors, undermining the reliability of significance claims in the small-effect regime where statistical testing is required the most.
By aligning the mean directions with a single Householder transformation, our approach targets dispersion differences more directly and yields better-calibrated permutation test $p$ values.
In addition, because permutation testing is well established in NLP but often considered computationally expensive in practice \cite{dror-etal-2018-hitchhikers}, we present a GPU-oriented implementation strategy that makes large-scale experimentation feasible.

\paragraph{Future Work}
Despite focusing on word-type token clouds, the proposed alignment and permutation framework applies to any two sets of representations, including sentence-embedding sets.
We did not evaluate sentence vectors in this study because of concerns that anisotropy and other geometric artifacts would complicate the interpretation of dispersion at the sentence level \cite{ethayarajh-2019-contextual}.
If future work mitigates these anisotropy concerns and enables reliable dispersion-based inference for sentence embeddings, the proposed test can be transferred to sets of prompts or questions used in LLM evaluation, providing a principled foundation for quantifying the \emph{breadth of knowledge} targeted by an evaluation suite independent of subjective human judgments.

Another possible direction is to apply the proposed framework to second-language vocabulary learning support \cite{ehara2025,ehara2022,ehara2012}. Among words with similar frequency, words with larger dispersion may be more difficult to learn, as they tend to appear across a wider range of contexts and senses.

\section*{Limitations}
Our empirical evaluation is necessarily limited in coverage. We tested a restricted set of corpora and model families; however, broader validation across genres, languages, and encoder architectures remains essential for assessing generality.
In particular, expanding the range of the corpora would clarify the robustness of the observed calibration improvements under different frequency profiles and contextual distributions.

A second limitation concerns the dispersion statistics themselves.
Similar to prior contextual diversity formulations based on von Mises-Fisher modeling, the underlying assumptions effectively treat the distribution of (normalized) token vectors for a word type as unimodal and isotropic, whereas real token clouds can be multimodal and/or anisotropic \cite{nagata-tanaka-ishii-2025-new}.
Such geometric deviations may affect both effect size and calibration of tests based on simplified distributional assumptions.
Prior large-scale evidence suggests that these assumptions can still yield observable and useful regularities at the word-type level, even when the true distributions are not perfectly isotropic \cite{nagata-tanaka-ishii-2025-new}. However, our work remains within this scope and does not claim to resolve anisotropy in general.

Finally, the practical constraints of pretrained tokenizers and vocabularies can limit the word types that can be analyzed cleanly (e.g., subword splitting), and this can introduce additional variability when comparing specific lexical items \cite{nagata-tanaka-ishii-2025-new}.

\section*{Ethical Considerations}

This study relies solely on publicly available pre-trained language models and existing linguistic resources such as WordNet. We do not collect, process, or generate any personal data, nor do we conduct experiments involving human subjects. The corpora used in our experiments (BNC, BCCWJ) are well-established linguistic resources obtained through appropriate academic licenses. Therefore, we do not foresee any significant ethical concerns arising from this work.

\section*{Acknowledgments}
This work was supported by JST PRESTO, Japan, Grant Number JPMJPR2363 and by JSPS KAKENHI Grant Number JP22K12287.
We are deeply grateful to the anonymous reviewers for their constructive feedback.

\bibliography{_custom,_extracted}

\begin{thebibliography}{26}
\providecommand{\natexlab}[1]{#1}

\bibitem[{Berg-Kirkpatrick et~al.(2012)Berg-Kirkpatrick, Burkett, and Klein}]{berg-kirkpatrick-etal-2012-empirical}
Taylor Berg-Kirkpatrick, David Burkett, and Dan Klein. 2012.
\newblock \href {https://aclanthology.org/D12-1091/} {An empirical investigation of statistical significance in {NLP}}.
\newblock In \emph{Proceedings of the 2012 Joint Conference on Empirical Methods in Natural Language Processing and Computational Natural Language Learning}, pages 995--1005, Jeju Island, Korea. Association for Computational Linguistics.

\bibitem[{{BNC Consortium}(2007)}]{bnc}
{BNC Consortium}. 2007.
\newblock \href {http://hdl.handle.net/20.500.14106/2554} {The {British National Corpus}, {XML} edition}.
\newblock License: http://www.natcorp.ox.ac.uk/docs/licence.html.

\bibitem[{Bond et~al.(2019)Bond, Janz, Maziarz, and Rudnicka}]{bond-etal-2019-testing}
Francis Bond, Arkadiusz Janz, Marek Maziarz, and Ewa Rudnicka. 2019.
\newblock \href {https://doi.org/10.18653/v1/2019.gwc-1.44} {Testing {Z}ipf{'}s meaning-frequency law with wordnets as sense inventories}.
\newblock In \emph{Proceedings of the 10th Global Wordnet Conference}, pages 342--352, Wroclaw, Poland. Global Wordnet Association.

\bibitem[{Devlin et~al.(2019)Devlin, Chang, Lee, and Toutanova}]{devlin-etal-2019-bert}
Jacob Devlin, Ming-Wei Chang, Kenton Lee, and Kristina Toutanova. 2019.
\newblock \href {https://doi.org/10.18653/v1/N19-1423} {{BERT}: Pre-training of deep bidirectional transformers for language understanding}.
\newblock In \emph{Proceedings of the 2019 Conference of the North {A}merican Chapter of the Association for Computational Linguistics: Human Language Technologies, Volume 1 (Long and Short Papers)}, pages 4171--4186, Minneapolis, Minnesota. Association for Computational Linguistics.

\bibitem[{Dror et~al.(2018)Dror, Baumer, Shlomov, and Reichart}]{dror-etal-2018-hitchhikers}
Rotem Dror, Gili Baumer, Segev Shlomov, and Roi Reichart. 2018.
\newblock \href {https://doi.org/10.18653/v1/P18-1128} {The hitchhiker{'}s guide to testing statistical significance in natural language processing}.
\newblock In \emph{Proceedings of the 56th Annual Meeting of the Association for Computational Linguistics (Volume 1: Long Papers)}, pages 1383--1392, Melbourne, Australia. Association for Computational Linguistics.

\bibitem[{Ehara(2022)}]{ehara2022}
Yo~Ehara. 2022.
\newblock \href {https://doi.org/10.1007/978-3-031-11644-5\_37} {An intelligent interactive support system for word usage learning in second languages}.
\newblock In \emph{Artificial Intelligence in Education - 23rd International Conference, {AIED} 2022, Durham, UK, July 27-31, 2022, Proceedings, Part {I}}, Lecture Notes in Computer Science, pages 453--464. Springer.

\bibitem[{Ehara(2025)}]{ehara2025}
Yo~Ehara. 2025.
\newblock \href {https://library.apsce.net/index.php/ICCE/article/view/5944} {Educational cone model in embedding vector spaces}.
\newblock In \emph{Proceedings of ICCE 2025: The 33rd International Conference on Computers in Education (short paper)}.

\bibitem[{Ehara et~al.(2012)Ehara, Sato, Oiwa, and Nakagawa}]{ehara2012}
Yo~Ehara, Issei Sato, Hidekazu Oiwa, and Hiroshi Nakagawa. 2012.
\newblock \href {https://aclanthology.org/C12-1049/} {Mining words in the minds of second language learners: Learner-specific word difficulty}.
\newblock In \emph{Proceedings of {COLING} 2012}, pages 799--814, Mumbai, India. The COLING 2012 Organizing Committee.

\bibitem[{Ethayarajh(2019)}]{ethayarajh-2019-contextual}
Kawin Ethayarajh. 2019.
\newblock \href {https://doi.org/10.18653/v1/D19-1006} {How contextual are contextualized word representations? {C}omparing the geometry of {BERT}, {ELM}o, and {GPT}-2 embeddings}.
\newblock In \emph{Proceedings of the 2019 Conference on Empirical Methods in Natural Language Processing and the 9th International Joint Conference on Natural Language Processing (EMNLP-IJCNLP)}, pages 55--65, Hong Kong, China. Association for Computational Linguistics.

\bibitem[{Giulianelli et~al.(2020)Giulianelli, Del~Tredici, and Fern{\'a}ndez}]{giulianelli-etal-2020-analysing}
Mario Giulianelli, Marco Del~Tredici, and Raquel Fern{\'a}ndez. 2020.
\newblock \href {https://doi.org/10.18653/v1/2020.acl-main.365} {Analysing lexical semantic change with contextualised word representations}.
\newblock In \emph{Proceedings of the 58th Annual Meeting of the Association for Computational Linguistics}, pages 3960--3973, Online.

\bibitem[{Graham et~al.(2014)Graham, Mathur, and Baldwin}]{graham-etal-2014-randomized}
Yvette Graham, Nitika Mathur, and Timothy Baldwin. 2014.
\newblock \href {https://doi.org/10.3115/v1/W14-3333} {Randomized significance tests in machine translation}.
\newblock In \emph{Proceedings of the Ninth Workshop on Statistical Machine Translation}, pages 266--274, Baltimore, Maryland, USA. Association for Computational Linguistics.

\bibitem[{Koehn(2004)}]{koehn-2004-statistical}
Philipp Koehn. 2004.
\newblock \href {https://aclanthology.org/W04-3250/} {Statistical significance tests for machine translation evaluation}.
\newblock In \emph{Proceedings of the 2004 Conference on Empirical Methods in Natural Language Processing}, pages 388--395, Barcelona, Spain. Association for Computational Linguistics.

\bibitem[{Kutuzov and Giulianelli(2020)}]{kutuzov-giulianelli-2020-uio}
Andrey Kutuzov and Mario Giulianelli. 2020.
\newblock \href {https://doi.org/10.18653/v1/2020.semeval-1.14} {{U}i{O}-{U}v{A} at {S}em{E}val-2020 task 1: Contextualised embeddings for lexical semantic change detection}.
\newblock In \emph{Proceedings of the Fourteenth Workshop on Semantic Evaluation}, pages 126--134, Barcelona (online). International Committee for Computational Linguistics.

\bibitem[{Maekawa et~al.(2014)Maekawa, Yamazaki, Ogiso, Maruyama, Ogura, Kashino, Koiso, Yamaguchi, Tanaka, and Den}]{bccwj}
Kikuo Maekawa, Makoto Yamazaki, Toshinobu Ogiso, Takehiko Maruyama, Hideki Ogura, Wakako Kashino, Hanae Koiso, Masaya Yamaguchi, Makiro Tanaka, and Yasuharu Den. 2014.
\newblock Balanced corpus of contemporary written {Japanese}.
\newblock \emph{Language Resources and Evaluation}, 48:345--371.

\bibitem[{Miller(1995)}]{wordnet}
George~A. Miller. 1995.
\newblock {WordNet}: A lexical database for {English}.
\newblock \emph{Communications of the ACM}, 38(11):39--41.

\bibitem[{Nagata and Tanaka-Ishii(2025)}]{nagata-tanaka-ishii-2025-new}
Ryo Nagata and Kumiko Tanaka-Ishii. 2025.
\newblock \href {https://doi.org/10.18653/v1/2025.acl-long.744} {A new formulation of {Z}ipf{'}s meaning-frequency law through contextual diversity}.
\newblock In \emph{Proceedings of the 63rd Annual Meeting of the Association for Computational Linguistics (Volume 1: Long Papers)}, pages 15323--15335, Vienna, Austria.

\bibitem[{Periti and Tahmasebi(2024)}]{periti-tahmasebi-2024-systematic}
Francesco Periti and Nina Tahmasebi. 2024.
\newblock \href {https://doi.org/10.18653/v1/2024.naacl-long.240} {A systematic comparison of contextualized word embeddings for lexical semantic change}.
\newblock In \emph{Proceedings of the 2024 Conference of the North American Chapter of the Association for Computational Linguistics: Human Language Technologies (Volume 1: Long Papers)}, pages 4262--4282, Mexico City, Mexico.

\bibitem[{Riezler and Maxwell(2005)}]{riezler-maxwell-2005-pitfalls}
Stefan Riezler and John~T. Maxwell. 2005.
\newblock \href {https://aclanthology.org/W05-0908/} {On some pitfalls in automatic evaluation and significance testing for {MT}}.
\newblock In \emph{Proceedings of the {ACL} Workshop on Intrinsic and Extrinsic Evaluation Measures for Machine Translation and/or Summarization}, pages 57--64, Ann Arbor, Michigan. Association for Computational Linguistics.

\bibitem[{Sch{\"o}nemann(1966)}]{Schonemann1966}
Peter~H. Sch{\"o}nemann. 1966.
\newblock \href {https://doi.org/10.1007/BF02289451} {A generalized solution of the orthogonal procrustes problem}.
\newblock \emph{Psychometrika}, 31(1):1--10.

\bibitem[{Vallebueno et~al.(2024)Vallebueno, Handan-Nader, Manning, and Ho}]{vallebueno-etal-2024-statistical}
Andrea Vallebueno, Cassandra Handan-Nader, Christopher~D Manning, and Daniel~E. Ho. 2024.
\newblock \href {https://doi.org/10.18653/v1/2024.emnlp-main.510} {Statistical uncertainty in word embeddings: {G}lo{V}e-{V}}.
\newblock In \emph{Proceedings of the 2024 Conference on Empirical Methods in Natural Language Processing}, pages 9032--9047, Miami, Florida, USA.

\bibitem[{van~der Maaten and Hinton(2008)}]{tsne}
Laurens van~der Maaten and Geoffrey Hinton. 2008.
\newblock Visualizing data using t-sne.
\newblock \emph{Journal of Machine Learning Research}, 9:2579--2605.

\bibitem[{Vuli{\'c} et~al.(2017)Vuli{\'c}, Gerz, Kiela, Hill, and Korhonen}]{vulic-etal-2017-hyperlex}
Ivan Vuli{\'c}, Daniela Gerz, Douwe Kiela, Felix Hill, and Anna Korhonen. 2017.
\newblock \href {https://doi.org/10.1162/COLI_a_00301} {{H}yper{L}ex: A large-scale evaluation of graded lexical entailment}.
\newblock \emph{Computational Linguistics}, 43(4):781--835.

\bibitem[{Warner et~al.(2025)Warner, Chaffin, Clavi{\'e}, Weller, Hallstr{\"o}m, Taghadouini, Gallagher, Biswas, Ladhak, Aarsen, Adams, Howard, and Poli}]{warner-etal-2025-smarter}
Benjamin Warner, Antoine Chaffin, Benjamin Clavi{\'e}, Orion Weller, Oskar Hallstr{\"o}m, Said Taghadouini, Alexis Gallagher, Raja Biswas, Faisal Ladhak, Tom Aarsen, Griffin~Thomas Adams, Jeremy Howard, and Iacopo Poli. 2025.
\newblock \href {https://doi.org/10.18653/v1/2025.acl-long.127} {Smarter, better, faster, longer: A modern bidirectional encoder for fast, memory efficient, and long context finetuning and inference}.
\newblock In \emph{Proceedings of the 63rd Annual Meeting of the Association for Computational Linguistics (Volume 1: Long Papers)}, pages 2526--2547, Vienna, Austria.

\bibitem[{Xypolopoulos et~al.(2021)Xypolopoulos, Tixier, and Vazirgiannis}]{xypolopoulos-etal-2021-unsupervised}
Christos Xypolopoulos, Antoine Tixier, and Michalis Vazirgiannis. 2021.
\newblock \href {https://doi.org/10.18653/v1/2021.eacl-main.297} {Unsupervised word polysemy quantification with multiresolution grids of contextual embeddings}.
\newblock In \emph{Proceedings of the 16th Conference of the European Chapter of the Association for Computational Linguistics: Main Volume}, pages 3391--3401, Online. Association for Computational Linguistics.

\bibitem[{Yamagiwa and Shimodaira(2025)}]{yamagiwa-shimodaira-2025-norm}
Hiroaki Yamagiwa and Hidetoshi Shimodaira. 2025.
\newblock \href {2025.coling-main.521/} {Norm of mean contextualized embeddings determines their variance}.
\newblock In \emph{Proceedings of the 31st International Conference on Computational Linguistics}, pages 7778--7808, Abu Dhabi, UAE.

\bibitem[{Zmigrod et~al.(2022)Zmigrod, Vieira, and Cotterell}]{zmigrod-etal-2022-exact}
Ran Zmigrod, Tim Vieira, and Ryan Cotterell. 2022.
\newblock \href {https://doi.org/10.18653/v1/2022.naacl-main.360} {Exact paired-permutation testing for structured test statistics}.
\newblock In \emph{Proceedings of the 2022 Conference of the North American Chapter of the Association for Computational Linguistics: Human Language Technologies}, pages 4894--4902, Seattle, United States.

\end{thebibliography}

\appendix

\section{Checklist Answers}
\label{sec:checklist}

This appendix provides answers to the ARR Responsible NLP Research checklist.
All section titles below correspond to the checklist item IDs.

\subsection{ID: A2 Potential Risks}

\noindent\textbf{Elaboration:}
Potential risks are discussed in the Limitations section.
Our method is a general-purpose statistical testing framework for comparing the semantic breadth of word embeddings.
Therefore, we do not foresee dual-use concerns or direct societal harms from this statistical methodology.



\subsection{ID: B1 Cite Creators Of Artifacts}
\noindent\textbf{Elaboration:}
As for corpora, we used the BNC \cite{bnc}, BCCWJ \cite{bccwj}, and WordNet \cite{wordnet}.

As for models we used:
BERT-tiny: \url{https://huggingface.co/prajjwal1/bert-tiny}.
BERT-large-uncased: \url{https://huggingface.co/google-bert/bert-large-uncased}.
BERT for Japanese, namely bert-large-japanese-v2: \url{https://huggingface.co/tohoku-nlp/bert-large-japanese-v2}.

\subsection{ID: B2 Discuss The License For Artifacts}
\noindent\textbf{Elaboration:}
We have confirmed that all corpora and models are distributed under licenses that permit their use for research purposes.

\subsection{ID: B3 Artifact Use Consistent With Intended Use}
\noindent\textbf{Elaboration:}
We have confirmed that all corpora and models are distributed under licenses that permit their use for research purposes.
Therefore, we consider their use in this study to be consistent with the intended use specified by their respective licenses.

\subsection{ID: B4 Data Contains Personally Identifying Information Or Offensive Content}

\noindent\textbf{Elaboration:}
We use only publicly available, well-established corpora (such as BNC and BCCWJ) and models that have been previously released for research purposes under their respective licenses.
We do not create new datasets or collect any personal data.
These corpora have undergone standard curation processes by their original creators.
Therefore, we do not consider that our paper contains personally identifying information or offensive content.

\subsection{ID: B5 Documentation of Artifacts and ID: B6 Statistics For Data}
\label{sec:b6}

\noindent\textbf{Elaboration:}
We documented basic information of the artifacts within the body texts of this paper. The more detailed documentation is available from the urls and papers in the artifact citations.

Table~\ref{tab:corpus_statistics} provides statistics for the corpora used in our experiments.

\begin{table}[h]
\centering
\begin{tabular}{lrl}
\hline
\textbf{Corpus} & \textbf{Size (tokens)} & \textbf{License} \\
\hline
BNC & 109,369,848 & BNC User Licence\textsuperscript{1} \\
BCCWJ & 124,102,859 & Academic License\textsuperscript{2} \\
\hline
\end{tabular}
\caption{Statistics of corpora used in experiments.}
\label{tab:corpus_statistics}
\end{table}

\noindent\textsuperscript{1}\url{http://www.natcorp.ox.ac.uk/docs/licence.html}\\
\textsuperscript{2}\url{https://clrd.ninjal.ac.jp/bccwj/en/index.html}

\noindent\textbf{Pretrained Models:}
\begin{itemize}
    \item \texttt{BERT-tiny}: 4.4M parameters (Apache 2.0)
    \item \texttt{BERT-large}: 340M parameters (Apache 2.0)
    \item \texttt{ModernBERT}: Apache 2.0
\end{itemize}

\noindent\textbf{Lexical Resource:}
\begin{itemize}
    \item WordNet: Used for sense count evaluation (WordNet License)
\end{itemize}


\subsection{ID: C1 Model Size And Budget}
\noindent\textbf{Elaboration:}
We conducted computational experiments to evaluate the proposed Householder-aligned permutation test:
\begin{itemize}
    \item \textbf{Main experiments (Section~4):} Type-I error and precision evaluation using WordNet sense counts
    \item \textbf{Additional experiments:} Type-I error control and statistical power across multiple corpus-model combinations
    \item \textbf{GPU acceleration:} Benchmark comparisons between CPU and GPU implementations
\end{itemize}

All models that we used in this paper are below 1 billion parameters.
The electricity fee for running two H100 GPUs are the budget for our computation.
For both BNC and BCCWJ, extracting all contextual embeddings for the target set of words took approximately 4 hours of GPU computation. The proposed GPU-based permutation tests take 20 minutes for 20,000 permutations of 300 words.

\subsection{ID: C2 Experimental Setup And Hyperparameters}

\noindent\textbf{Elaboration:}
All experimental setups and hyperparameters are described in Appendix A and Section 4.



Our experimental setup uses the following configurations:

\noindent\textbf{Main Experiments (Section~4):}
\begin{table}[h]
\centering
\begin{tabular}{ll}
\hline
\textbf{Parameter} & \textbf{Value} \\
\hline
Corpus & BNC \\
Model & ModernBERT \\
Number of word pairs & 300 per gap setting \\
Dispersion ranking gap & 1--10 \\
Significance level ($\alpha$) & 0.01 \\
Gold standard & WordNet synset counts \\
\hline
\end{tabular}
\end{table}

\vspace{0.5em}
\noindent\textbf{Additional Experiments:}
\begin{table}[h]
\centering
\begin{tabular}{p{3cm}l}
\hline
\textbf{Parameter} & \textbf{Value} \\
\hline
Corpus--Model combinations & BNC + BERT-tiny \\
 & BNC + ModernBERT \\
 & BCCWJ + BERT-large \\
Number of word pairs & 500 \\
Number of permutations ($B$) & 5,000 \\
Significance level ($\alpha$) & 0.05 \\
Gap conditions & 50 (Type-I error) \\
 & 100 (Statistical power) \\
\hline
\end{tabular}
\end{table}

\noindent\textbf{Algorithm Parameters:}
\begin{itemize}
    \item Embedding normalization: $\ell_2$-normalization to unit sphere
    \item Dispersion statistic: Mean resultant length (MRL)
    \item Test statistic: Log-volume difference (Eq.~11)
    \item Alignment method: Householder reflection (single transformation)
\end{itemize}

\subsection{ID: C3 Descriptive Statistics}

\noindent\textbf{Elaboration:}
We report comprehensive descriptive statistics for our experimental results:

\begin{itemize}
    \item \textbf{Table~1:} Type-I Error and Precision comparison at dispersion ranking gaps 1--10
    \item \textbf{Table~2:} Rejection rates under null (gap=50) and alternative (gap=100) hypotheses for three corpus--model combinations
    \item \textbf{Figure~2:} Visualization of Type-I Error rate and Precision across gap values
\end{itemize}

\noindent\textbf{Key Results Summary:}
\begin{itemize}
    \item Type-I error reduction: 32.5\% compared to baseline (naive permutation test)
    \item GPU speedup: 23$\times$ over CPU baseline
    \item At gap=10: Type-I Error reduced from 4.0\% to 2.67\%; Precision improved from 41.7\% to 62.5\%
\end{itemize}

\subsection{ID: C4 Parameters For Packages}

\noindent\textbf{Elaboration:}
We used PyTorch \url{https://pytorch.org/}, HuggingFace \url{https://huggingface.co/}
models.
All parameters are described in the experiment setup descriptions of Section 4 and this Appendix.

\subsection{ID: E1 Information About Use Of AI Assistants}

\noindent\textbf{Elaboration:}
We used codex CLI and Claude code to implement our proposed methods. All code is manually checked.
We used ChatGPT and Claude to improve the English of the paper.

\section{Algorithm Description}
This appendix presents the algorithms used in our method: \algref{alg:householder_perm}  describes the basic Householder-aligned permutation test, and \algref{alg:gpu_perm} gives its GPU-oriented implementation.

\begin{algorithm}[t!]
\caption{Householder-Aligned Permutation Test}
\label{alg:householder_perm}
\begin{algorithmic}[1]
\Require Unit-normalized embeddings $X=\{\mathbf{x}_i\}_{i=1}^{n}$, $Y=\{\mathbf{y}_j\}_{j=1}^{m}$; number of permutations $B$; significance level $\alpha$
\Ensure $p$-value for testing $H_0$: $\text{disp}(X)=\text{disp}(Y)$

\Statex \textbf{// Step 1: Compute sample mean directions}
\State $\bar{\mathbf{x}} \gets \frac{1}{n}\sum_{i=1}^n \mathbf{x}_i$; \quad $\bar{\mathbf{y}} \gets \frac{1}{m}\sum_{j=1}^m \mathbf{y}_j$
\State $\hat{\boldsymbol{\mu}}_x \gets \bar{\mathbf{x}}/\|\bar{\mathbf{x}}\|_2$; \quad $\hat{\boldsymbol{\mu}}_y \gets \bar{\mathbf{y}}/\|\bar{\mathbf{y}}\|_2$

\Statex \textbf{// Step 2: Construct Householder reflection}
\State $\mathbf{u} \gets (\hat{\boldsymbol{\mu}}_x - \hat{\boldsymbol{\mu}}_y)/\|\hat{\boldsymbol{\mu}}_x - \hat{\boldsymbol{\mu}}_y\|_2$

\Statex \textbf{// Step 3: Align $X$ to $Y$'s mean direction}
\For{$i = 1$ to $n$}
    \State $\mathbf{x}'_i \gets \mathbf{x}_i - 2\mathbf{u}(\mathbf{u}^\top\mathbf{x}_i)$ \Comment{Eq.~\eqref{eq:householder_fast}}
\EndFor
\State $X' \gets \{\mathbf{x}'_i\}_{i=1}^{n}$

\Statex \textbf{// Step 4: Compute observed test statistic}
\State $r_{X'} \gets \left\|\frac{1}{n}\sum_{i=1}^n \mathbf{x}'_i\right\|_2$; \quad $r_Y \gets \left\|\frac{1}{m}\sum_{j=1}^m \mathbf{y}_j\right\|_2$
\State $T_{\mathrm{obs}} \gets \log(1/g_d(r_{X'})) - \log(1/g_d(r_Y))$ \Comment{Eq.~\eqref{eq:tobs}}

\Statex \textbf{// Step 5: Permutation test on aligned data}
\State $Z \gets X' \cup Y$ \Comment{Pool aligned samples}
\State $\text{count} \gets 0$
\For{$b = 1$ to $B$}
    \State Randomly partition $Z$ into $(X^{(b)}, Y^{(b)})$ with sizes $(n, m)$
    \State $r_1^{(b)} \gets \left\|\frac{1}{n}\sum_{\mathbf{z} \in X^{(b)}} \mathbf{z}\right\|_2$; \quad $r_2^{(b)} \gets \left\|\frac{1}{m}\sum_{\mathbf{z} \in Y^{(b)}} \mathbf{z}\right\|_2$
    \State $T^{(b)} \gets \log(1/g_d(r_1^{(b)})) - \log(1/g_d(r_2^{(b)}))$
    \If{$T^{(b)} \ge T_{\mathrm{obs}}$}
        \State $\text{count} \gets \text{count} + 1$
    \EndIf
\EndFor

\Statex \textbf{// Step 6: Compute $p$-value}
\State $p \gets (1 + \text{count}) / (B + 1)$ \Comment{Eq.~\eqref{eq:pvalue}}
\State \Return $p$
\end{algorithmic}
\end{algorithm}


\begin{algorithm}[t!]
\caption{GPU-Accelerated Permutation Statistics}
\label{alg:gpu_perm}
\begin{algorithmic}[1]
\Require Aligned pooled matrix $\mathbf{X}\in\mathbb{R}^{N\times d}$ (rows of $Z=X'\cup Y$); group sizes $(n,m)$ with $N=n+m$; number of permutations $B$; block size $B_0$
\Ensure Exceedance count for $p$-value computation

\Statex \textbf{// Precompute total sum (once)}
\State $\mathbf{t} \gets \mathbf{1}^\top\mathbf{X} \in \mathbb{R}^{d}$ \Comment{Sum of all rows}
\State $\text{count} \gets 0$

\Statex \textbf{// Process permutations in GPU-friendly blocks}
\For{each block $k = 1, 2, \ldots, \lceil B/B_0 \rceil$}
    \Statex \quad \textbf{// Generate sign matrix for this block}
    \State $\mathbf{S}_{\text{blk}} \gets$ random $\{+1,-1\}^{B_0\times N}$ with exactly $n$ entries $+1$ per row
    
    \Statex \quad \textbf{// Single GEMM call for all permutations in block}
    \State $\mathbf{U}_{\text{blk}} \gets \mathbf{S}_{\text{blk}}\mathbf{X} \in \mathbb{R}^{B_0 \times d}$ \Comment{Eq.~\eqref{eq:gemm}}
    
    \Statex \quad \textbf{// Recover group sums via broadcasting}
    \For{$b = 1$ to $B_0$ \textbf{in parallel}}
        \State $\boldsymbol{\sigma}^{(b)}_1 \gets (\mathbf{t} + \mathbf{U}_{\text{blk},b,:})/2$
        \State $\boldsymbol{\sigma}^{(b)}_2 \gets (\mathbf{t} - \mathbf{U}_{\text{blk},b,:})/2$
        
        \Statex \quad\quad \textbf{// Compute mean resultant lengths}
        \State $r^{(b)}_1 \gets \|\boldsymbol{\sigma}^{(b)}_1/n\|_2$; \quad $r^{(b)}_2 \gets \|\boldsymbol{\sigma}^{(b)}_2/m\|_2$
        
        \Statex \quad\quad \textbf{// Compute test statistic}
        \State $T^{(b)} \gets \log(1/g_d(r^{(b)}_1)) - \log(1/g_d(r^{(b)}_2))$
    \EndFor
    
    \Statex \quad \textbf{// Update exceedance count (streaming)}
    \State $\text{count} \gets \text{count} + \sum_{b=1}^{B_0} \mathbb{I}[T^{(b)} \ge T_{\mathrm{obs}}]$
\EndFor

\State \Return count
\end{algorithmic}
\end{algorithm}

\begin{table*}[t!]
\centering
\small
\begin{tabular}{l r r r r r r}
\hline
Word & Total & Baseline / greater & Proposed / greater & Baseline / two-sided & Proposed / two-sided \\
\hline
ablaze        & 160 & 0.2441 & 0.2474 & 0.4895 & 0.4956 \\
accolade      & 160 & 0.6356 & 0.6345 & 0.7330 & 0.7356 \\
actuality     & 156 & 0.7650 & 0.7644 & 0.4702 & 0.4714 \\
acheson       & 154 & 0.4623 & 0.4613 & 0.9205 & 0.9191 \\
adaptability  & 153 & 0.5581 & 0.5578 & 0.8827 & 0.8841 \\
adenauer      & 152 & 0.0369 & 0.0342 & 0.0756 & 0.0709 \\
additive      & 151 & 0.5910 & 0.5918 & 0.8148 & 0.8139 \\
absentee      & 150 & 0.6509 & 0.6534 & 0.6997 & 0.6950 \\
abnormally    & 149 & 0.3397 & 0.3412 & 0.6888 & 0.6901 \\
abstinence    & 147 & 0.5419 & 0.5395 & 0.9168 & 0.9186 \\
aberration    & 145 & 0.9040 & 0.9056 & 0.1845 & 0.1815 \\
acetate       & 145 & 0.6496 & 0.6489 & 0.7017 & 0.7019 \\
abstracted    & 139 & 0.2506 & 0.2562 & 0.5095 & 0.5173 \\
abyss         & 137 & 0.1744 & 0.1724 & 0.3501 & 0.3466 \\
abatement     & 136 & 0.4426 & 0.4416 & 0.8877 & 0.8860 \\
according     & 135 & 0.3434 & 0.3443 & 0.6827 & 0.6846 \\
acrimonious   & 135 & 0.9729 & 0.9628 & 0.0519 & 0.0725 \\
abstain       & 132 & 0.0420 & 0.0363 & 0.0841 & 0.0734 \\
accomplishment& 130 & 0.4508 & 0.4498 & 0.8928 & 0.8919 \\
acne          & 130 & 0.7160 & 0.7159 & 0.5633 & 0.5644 \\
\hline
\end{tabular}
\caption{Same-word split-half sanity-check p-values for 20 English words. ``Baseline'' denotes the permutation test without Householder alignment, and ``Proposed'' denotes the Householder-aligned permutation test.}
\label{tab:sanity_split_half_pvalues}
\end{table*}

\section{Qualitative Examples of Statistical Testing}
To make the calibration-power trade-off in Table~\ref{tab:rejection_rates} more concrete (Japanese; BCCWJ/BERT-large),
we inspected the pairs whose significance changed after alignment and presented representative examples.

Baseline-only rejections (rejected by the baseline but not by the proposed method) include several
content-word pairs, e.g., \jp{普段}--\jp{毎年} (\textit{fudan}, usually / in everyday situations -- \textit{maitoshi}, every year),
\jp{書物}--\jp{気管} (\textit{shomotsu}, books / written works -- \textit{kikan}, trachea / airway), and
\jp{家電}--\jp{騒音} (\textit{kaden}, home appliances -- \textit{souon}, noise / noise pollution). In contrast, many pairs that remain
statistically significant under both methods involve verb-/function-like tokens versus content nouns,
e.g., \jp{避け}--\jp{最低} (\textit{sake}, avoid(ing) -- \textit{saitei}, the worst / lowest) and
\jp{叫ん}--\jp{サラリーマン} (\textit{saken}, shout(ing) -- \textit{sarari-man}, office worker / salaryman)
(we also observe a few noun--noun cases such as \jp{仏教}--\jp{官僚}
(\textit{bukkyo}, Buddhism -- \textit{kanryou}, bureaucrat(s)) and \jp{下着}--\jp{村人} (\textit{shitagi}, underwear -- \textit{murabito}, villager(s))).

\begin{figure}[t!]
\centering
\includegraphics[width=1.0\columnwidth]{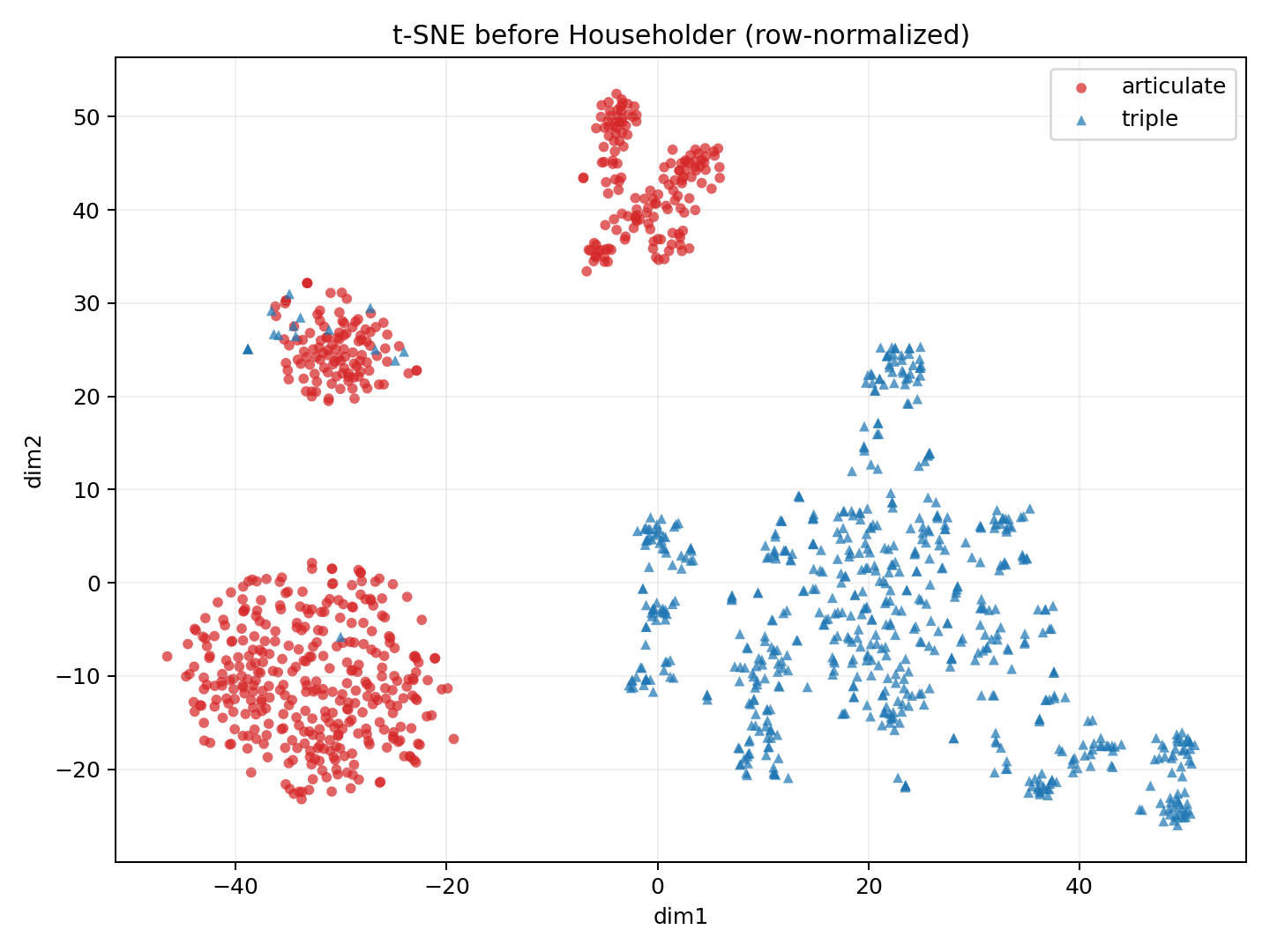}
\caption{Baseline, i.e., t-SNE visualization before applying Householder transform.}
\label{fig:tsne_before}
\end{figure}
\begin{figure}[t!]
\centering
\includegraphics[width=1.0\columnwidth]{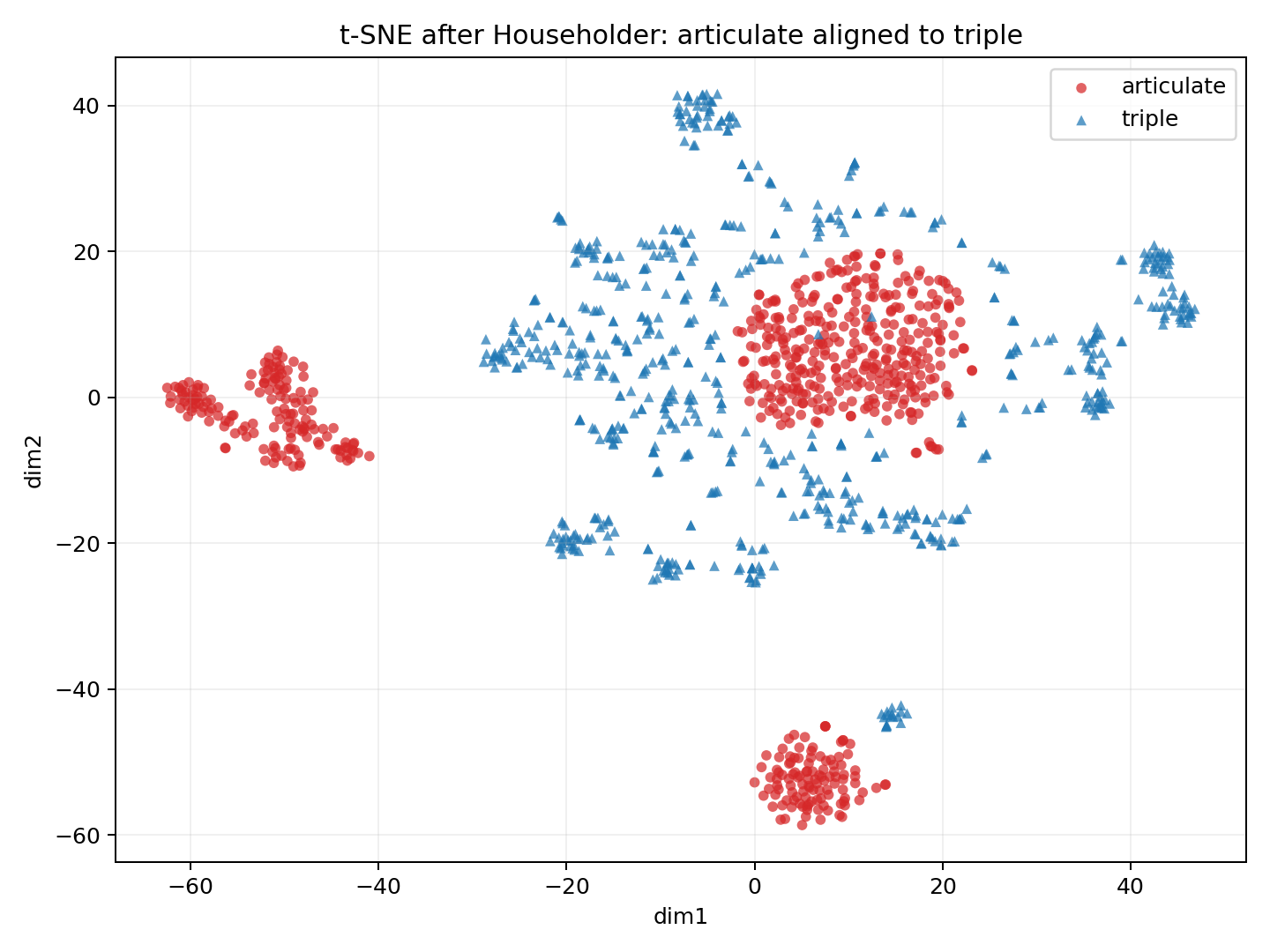}
\caption{Proposed, i.e., t-SNE visualization after applying Householder transform.}
\label{fig:tsne_after}
\end{figure}

\section{Proof Sketch} \label{sec:proofsketch}
In this section, we provide a proof sketch that the transformation $\mathbf{R}$ that maximizes the MRL of the merged set is given by the Householder transform.
To avoid display/rendering issues, we write the sums in a simplified form.

The MRL for the merged set can be written as
\begin{align}
\mathrm{MRL}
&= \frac{1}{n+m} \left\| \sum_i \mathbf{R}\mathbf{x}_i + \sum_j \mathbf{y}_j \right\| \\
&= \frac{1}{n+m} \left\| \mathbf{R} \sum_i \mathbf{x}_i + \sum_j \mathbf{y}_j \right\| \\
&= \frac{1}{n+m} \left\| n\,\mathbf{R}\bar{\mathbf{x}} + m\,\bar{\mathbf{y}} \right\|.
\end{align}

Thus, the MRL reduces to the norm of a sum of two vectors.
By the triangle inequality, this quantity is maximized if and only if the two vectors are parallel and not in opposite directions:
\begin{align}
\left\| n\,\mathbf{R}\bar{\mathbf{x}} + m\,\bar{\mathbf{y}} \right\|
\le
\left\| n\,\mathbf{R}\bar{\mathbf{x}} \right\|
+
\left\| m\,\bar{\mathbf{y}} \right\| \nonumber \\
=
n \left\| \mathbf{R}\bar{\mathbf{x}} \right\|
+
m \left\| \bar{\mathbf{y}} \right\|,
\end{align}
if and only if $\mathbf{R}\bar{\mathbf{x}}$ and $\bar{\mathbf{y}}$ are parallel,
i.e., $\mathbf{R}\bar{\mathbf{x}} = C \bar{\mathbf{y}}$ where $C > 0$.

In particular, when $\mathbf{R} = \mathbf{H}$, as  $\mathbf{H}\hat{\boldsymbol{\mu}}_x=\hat{\boldsymbol{\mu}}_y$,  $\hat{\boldsymbol{\mu}}_x=\bar{\mathbf{x}}/\|\bar{\mathbf{x}}\|$, and
$\hat{\boldsymbol{\mu}}_y=\bar{\mathbf{y}}/\|\bar{\mathbf{y}}\|$, we have
\begin{align}
\mathbf{H}\bar{\mathbf{x}} / \|\bar{\mathbf{x}}\|
=
\bar{\mathbf{y}} / \|\bar{\mathbf{y}}\|.
\end{align}

Therefore, the maximum MRL of the merged set is achieved by rotating $\mathbf{X}$
using the Householder transform.
Note that the MRL is upper-bounded by
\begin{align}
\frac{1}{n+m}
\left(
n \|\mathbf{R}\bar{\mathbf{x}}\|
+
m \|\bar{\mathbf{y}}\|
\right) \nonumber \\
=
\frac{1}{n+m}
\left(
n \|\bar{\mathbf{x}}\|
+
m \|\bar{\mathbf{y}}\|
\right),
\end{align}
since $\mathbf{R}$ is orthogonal.

\section{Sanity Check} \label{sec:sanity_check}

In this section, we provide an example of sanity check experiments that we explained in Section~\ref{sec:sanitycheckhonbun}.
As shown in Table~\ref{tab:sanity_split_half_pvalues}, when occurrences of the same word are randomly divided into two groups, the two groups already have very similar mean vectors even without alignment. As a result, the baseline and the proposed method yield almost identical p-values in this same-word null setting, indicating that little difference in performance is expected between the two methods in such cases.

\section{Qualitative visualization} \label{sec:tsne}

In this section, we revisit how the Householder transform aligns the mean vectors of the contextualized word-embedding sets for the two words.
Figure~\ref{fig:tsne_before} and Figure~\ref{fig:tsne_after} show a qualitative t-SNE visualization \cite{tsne} for the word pair \textit{articulate}-\textit{triple} before and after Householder alignment. Before alignment, the two token clouds occupy more clearly separated regions in this 2D visualization. After Householder alignment, the two sets appear more mixed, which is consistent with the role of the proposed method in reducing nuisance mean-direction mismatch before permutation. We emphasize, however, that this figure is provided only for intuition: low-dimensional projections such as t-SNE do not faithfully preserve the original high-dimensional geometry or dispersion statistics, and should not be treated as primary evidence.

\end{document}